\documentclass[a4paper]{article}

\usepackage{INTERSPEECH_v2}
\usepackage{multirow}

\title{ASR error management for improving spoken language understanding}
\name{Edwin Simonnet$^1$, Sahar Ghannay$^1$, Nathalie Camelin$^1$, Yannick Est\`eve$^1$, Renato De Mori$^{2,3}$}
\address{
  $^1$LIUM, Le Mans University, France\\
  $^2$LIA, University of Avignon, France\\
  $^3$McGill University, Montreal, Canada}

\email{firstname.lastname@univ-lemans.fr, rdemori@cs.mcgill.ca}

\begin{document}

\maketitle
\begin{abstract}
This paper addresses the problem of automatic speech recognition (ASR) error detection and their use for improving spoken language understanding (SLU) systems. 
In this study, the SLU task consists in automatically extracting, from ASR transcriptions, semantic concepts and concept/values pairs in a \textsl{e.g} touristic information system. 
An approach is proposed for enriching the set of semantic labels with error specific labels and by using a recently proposed neural approach based on word embeddings to compute well calibrated ASR confidence measures. Experimental results are reported showing that it is possible to decrease significantly the Concept/Value Error Rate with a state of the art system, outperforming previously published results performance on the same experimental data.
It also shown that combining an SLU approach based on conditional random fields with a neural encoder/decoder attention based architecture, it is possible to effectively identifying confidence islands and uncertain semantic output segments useful for deciding appropriate error handling actions by  the dialogue manager strategy.

\end{abstract}

\section{Introduction}
In spite of impressive research efforts and recent results, systems for semantic interpretation of text and speech still make errors. Some of the problems common to text and speech are: difficulty of concept mention localization, ambiguities intrinsic in localized mentions, deficiency to identify sufficient contextual constraints for solving interpretation ambiguities.
Additional problems are introduced by the interaction between a spoken language understanding (SLU) system and an error prone automatic speech recognition (ASR) system. 
ASR errors may affect the mention of a concept, the value of a concept instance. Furthermore, the hypothesization of concepts and values depends, among other things, on the context in which their mention is localized. Thus, context errors may also introduce errors in concept mention location and hypothesization. 


The focus of this paper\footnote{Thanks to the ANR agency for funding through the CHIST-ERA ERA-Net JOKER under the con- tract number ANR-13-CHR2-0003-05.} is on the introduction of suitable ASR confidence measures for localizing ASR word errors that may affect SLU performance. They are used as additional SLU features to be combined with lexical and syntactic features useful for characterizing concept mentions. For this purpose, an  ASR error detection sub-system has been endowed with confidence features based on syntactic dependencies and other semantically relevant word features. Two SLU architectures equipped with other sets of confidence and specific word features are introduced. The architectures are based on conditional random fields (CRF) and an encoder-decoder neural network structure with a mechanism of attention (NN-EDA).
Experimental results showing significant reduction for the French MEDIA corpus on concepts and concept value pairs confirm the expected benefit of introducing semantic specific ASR features. Optimal combinations of these architectures provide additional improvements with a concept error rate (CER) relative reduction of 18.9\% and a concept-value error rate (CVER) relative reduction of 10.3\% with respect to a baseline described in \cite{hahn2011comparing} not using these features and based only on CRFs.


\section{Related work}

SLU systems are error prone. Part of them are caused by certain types of ASR errors. 
In general, ASR errors are reduced by estimating model parameters by minimizing the expected word error rate \cite{mangu2000finding}. The effect of word errors can be controlled by associating a single sentence hypothesis with word confidence measures. In \cite{yu2011calibration} methods are proposed for constructing confidence features for improving the quality of a semantic confidence measure. Methods proposed for confidence calibration are based on the maximum entropy model with distribution constraints, the conventional artificial neural network, and the deep belief network (DBN). The latter two methods show slightly superior performance but higher computational complexity compared to the first one. 
More recently \cite{ogawa2015asr}, new features and bidirectional recurrent neural networks (RNN) have been proposed for ASR error detection. 
Most SLU systems reviewed in \cite{tur2011spoken} generate hypotheses of semantic frame slot tags expressed in a spoken sentence analyzed by an ASR system. The use of deep neural networks (DNN) appeared in more recent systems as described in \cite{mesnil2015using}.
Bidirectional RNNs with long-short term memory (LSTM) have been used for semantic frame slot tagging \cite{hakkani2016multi}. In \cite{reddy2016transforming}, LSTMs have been proposed with a mechanism of attention for parsing text sentences to logical forms.
Following \cite{ma2015dependency}, in \cite{chen2016syntax} a convolutional neural network (CNN) is proposed for encoding the representation of knowledge expressed in a spoken sentence. This encoding is used as an attention mechanism for constraining the hypothesization of slot tags expressed in the same sentence. 
Most recent papers using sophisticated SLU architectures based on RNNs have the best sequence of word hypotheses as input passed by an ASR system. In this paper, two SLU architectures are considered. The first one, based an encoder with bidirectional gated recurrent units (GRU) used for machine translation \cite{simonnet2015ma}, integrates context information with an attention based decoder as in \cite{Cho2014}. The second one integrates context information in the same architecture used in \cite{hahn2011comparing} based on conditional random fields (CRF). 
Both SLU systems receive word hypotheses generated by the same ASR sub-system and scored with confidence measures computed by a neural architecture with new types of embeddings and semantically relevant confidence features.

\section{ASR error detection and confidence measure}
\label{sec:ASR}

Two different confidence measures are used for error detection. The first one is the word \textit{posterior} probability computed with confusion networks as described in \cite{mangu2000finding}. The other one is a variant of a new  approach, introduced in \cite{ghannay2015word,ghannay2016acoustic}. The latter measure is computed with a Multi-Stream Multi-Layer Perceptron (MS-MLP) architecture, fed by different heterogeneous confidence features. Among them, the most relevant for SLU are word embeddings of the targeted word and its neighbors, length of the current word, language model backoff behavior, part of speech (POS) tags, syntactic dependency labels and word governors. Other features, such as prosodic features and acoustic word embeddings described in \cite{ghannay2015combining} and \cite{ghannay2016acoustic} could also be used but were not considered in the experiments described in this paper.  A particular attention was carried on the word embeddings computation, which is the result of a combination of different well known word embeddings (CBOW, Skip-gram, GloVe) made through the use of a neural auto-encoder in order to improve the performances of this ASR error detection system \cite{ghannay2016word}. 

The MS-MLP proposed here for ASR error detection has two output units. They compute scores for \textsl{Correct} and \textsl{Error} labels associated with an ASR generated hypothesis. This hypothesis is evaluated by the softmax value of the \textsl{Correct} label scored with the MS-MLP.
Experiments have shown that this is a calibrated confidence measure more effective than word posterior probability when comparison is based on the Normalized Cross Entropy (NCE) \cite{ghannay2015combining}, which measures the information contribution provided by confidence knowledge.

Table~\ref{tab:MC_NCE} shows the NCE values obtained by these two confidence measures on the MEDIA test data whose details can be found in section~\ref{sec:exp}.

\begin{table}[hb]
\vspace{-0.1cm}
\centering
\begin{tabular}{|l|c|c|}
  \hline
   & PAP & MS-MLP \\
  \hline
   NCE & 0.147 & 0.462 \\
  \hline
\end{tabular}
\caption{\label{tab:MC_NCE} Comparison of ASR error prediction capabilities of probability \textsl{a posteriori} and confidence measure derived from the MS-MLP ASR error detection system in terms of Normalized Cross Entropy on the MEDIA test set.}
\vspace{-0.2cm}
\end{table}

\begin{figure}[!htbp]
\includegraphics [width=\columnwidth]{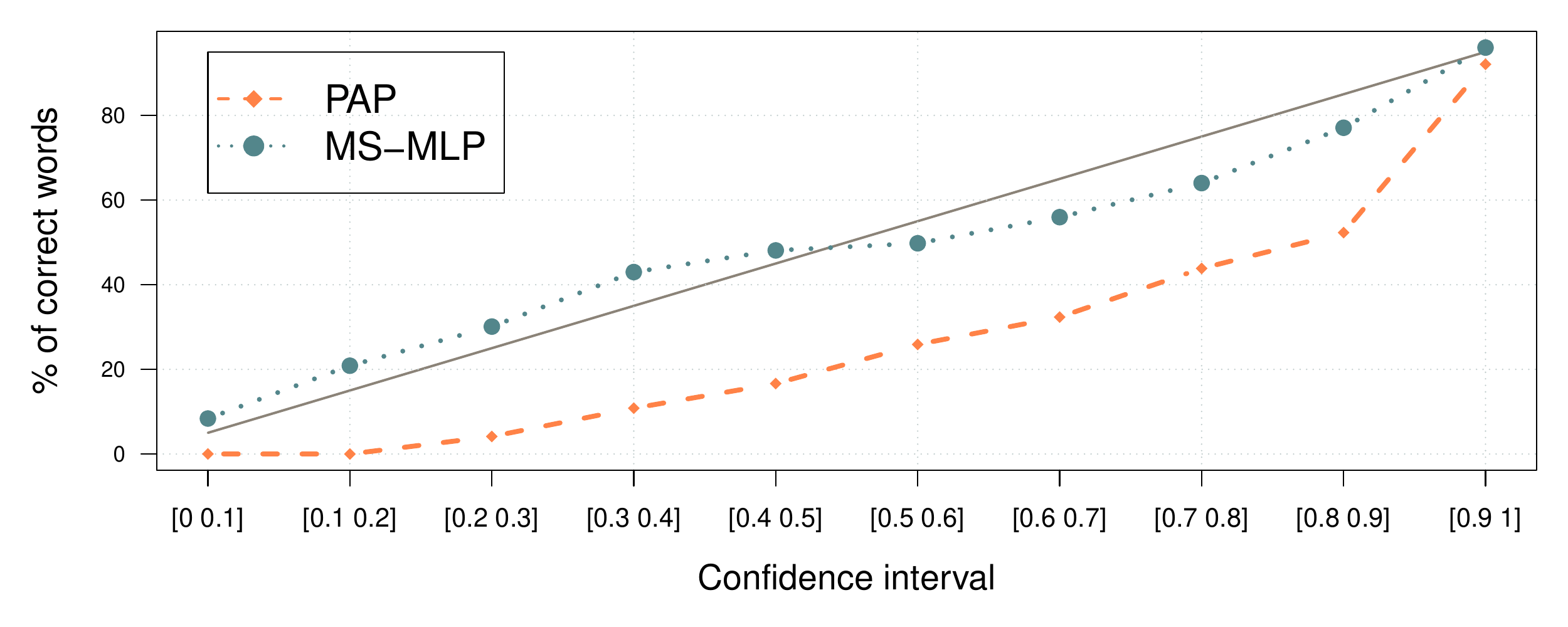} 
\vspace{-0.8cm}
\caption{\label{fig:MC_calibration} Comparison of ASR error prediction capabilities of probability \textsl{a posteriori} and confidence measure derived from the MS-MLP ASR error detection system on the MEDIA test set.}
\end{figure}

\vspace{-0.3cm}
Figure~\ref{fig:MC_calibration} shows the predictive capability of the confidence measure based on MLP-MS compared to word posterior probability on the MEDIA test data. The curve shows the predicted percentage of correct words as a function of confidence intervals. 
The best measure is the one for which percentages are the closest to the diagonal line.


\vspace{-0.2cm}

Thanks to these confidences measures, we expect to get relevant information in order to better handle ASR errors in a spoken language understanding framework.



\section{SLU features and architectures}

Two basis SLU architectures are considered to carry experiments on the MEDIA corpus (described in sub-section \ref{sec:media}). The first one is an encoder/decoder recurrent neural architecture with a mechanism of attention (NN-EDA) similar to the one used for machine translation proposed in \cite{Cho2014}. The second one is based on conditional random fields (CRF - \cite{lafferty2001conditional}). Both architectures  build their training model on the same features encoded with continuous values in the first one and discrete values in the second one

\subsection{Set of Features}
Word features, including those defined for facilitating the association of a word with a semantic content, are defined as follows:
\begin{itemize} 
\vspace{-0.2cm}
\item the word itself 
\vspace{-0.2cm}
\item its pre-defined semantic categories which belongs to: 
\begin{itemize}
\vspace{-0.2cm}
\item MEDIA specific categories: like names of the streets, cities or hotels, lists of room equipments, food type, \ldots ~\textit{e.g.}: TOWN for Paris
\vspace{-0.2cm}
\item more general categories: like figures, days, months, \ldots ~\textit{e.g.}: FIGURE for thirty-three. 
\end{itemize} 
\vspace{-0.2cm}
\item a set of syntactic features: the MACAON tool \cite{macaon2010} is applied to the whole turn in order to obtain for each word its following tags:  the lemma, the POS tag, its word governor and its relation with the current word. 
\vspace{-0.2cm}
\item a set of morphological features: the 1-to-4 first letter ngrams, the 1-to-4 letter last ngrams of the word and a binary feature that indicates if the first letter is an upper one. 
\vspace{-0.2cm}
\item the two ASR confidence measures : the ASR \textit{posterior} probability (\textit{pap}) and the MS-MLP confidence measure as described in section \ref{sec:ASR}.
\end{itemize}

The two SLU architectures take all those features except the two confidence measures that can be taken partially: one or another or both according to the most powerful configuration. The SLU architectures also need to be calibrated on their respective hyper-parameters in order to give the best results. The way the best configuration is chosen is described in \ref{ssec:results}.


\subsection{Neural EDA system} 

\begin{figure}[!htbp]
\includegraphics [width=1.1\columnwidth]{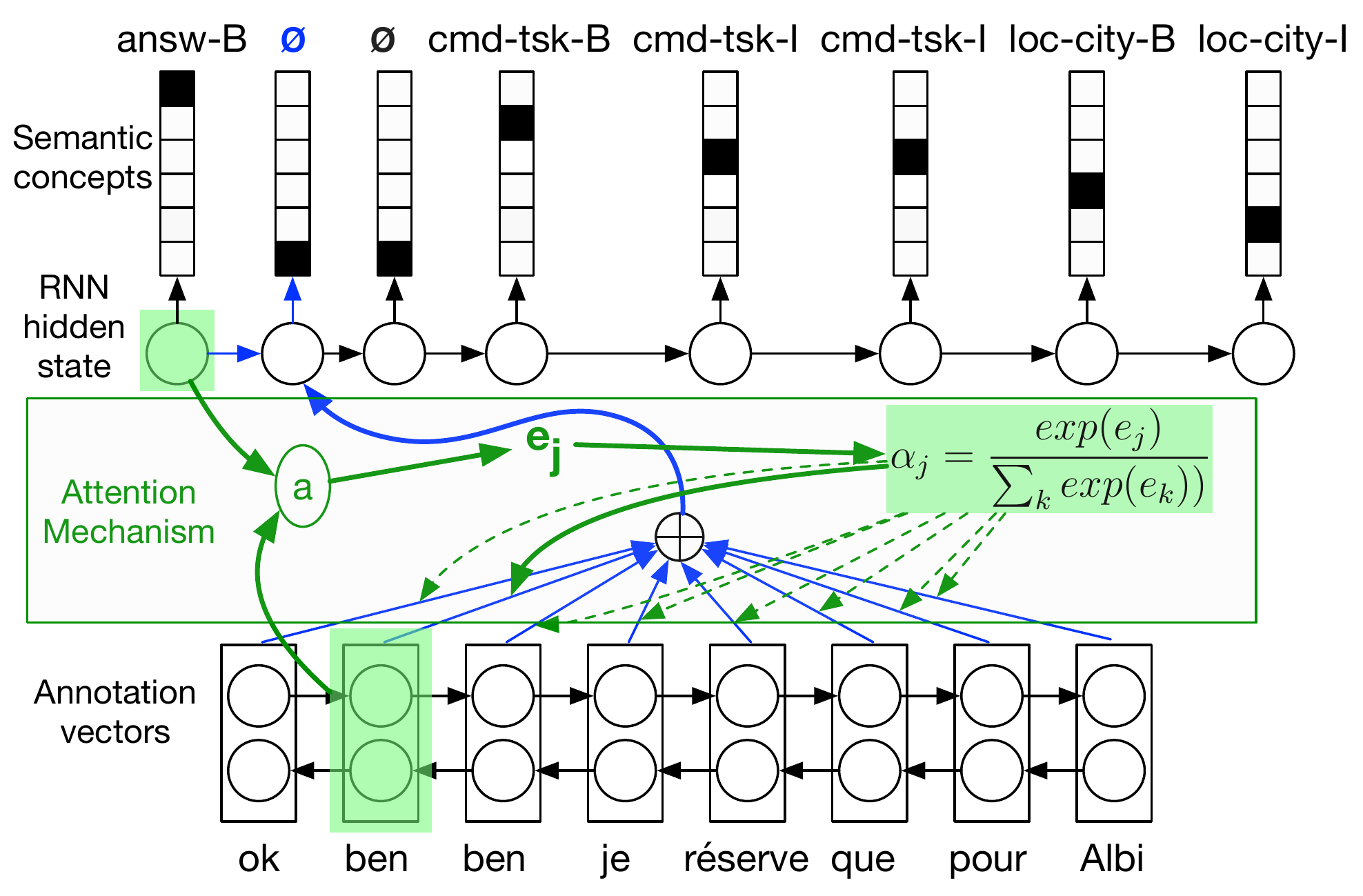} 
\caption{\label{fig:nn_eda} Architecture of the NN-EDA SLU system.}
\end{figure}

The proposed RNN encoder-decoder architecture with an attention-based mechanism (NN-EDA) is inspired from a machine translation architecture and depicted in figure~\ref{fig:nn_eda}. The concept tagging process is considered as a translation problem from words (source language) to semantic concept tags (target language). 
This bidirectional RNN encoder is based on Gated Recurrent Units (GRU) and computes an annotation $h_i$ for each word $w_i$ from the input sequence {$w_1$ , ... ,$w_I$ }. 
This annotation is the concatenation of the matching forward hidden layer state and the backward hidden layer state obtained respectively by the forward RNN and the backward RNN comprising the bidirectional RNN. 
Each annotation contains the summaries of the dialogue turn contexts respectively preceding and the following a considered word.


The sequence of annotations {$h_1$ , ... ,$h_I$ } is used by the decoder to compute a context vector $c_t$ (represented as a circle with a cross in figure~\ref{fig:nn_eda}). A context vector is recomputed after each emission of an output label. This computation takes into account a weighted sum of all the annotations computed by the encoder. This weighting depends on the current output target, and is the core of the attention mechanism: a good estimation of these weights $\alpha_{tj}$  allows the decoder to choose parts of the input sequence to pay attention to. This context vector is used by the decoder in conjunction with the previous emitted label output $y_{t-1}$ and the current state $s_t$ of the hidden layer of a RNN to make a decision about the current label output $y_t$.
A more detailled description of recurent neural networks and attention based ones can be found in \cite{simonnet2015ma}.

\vspace{-0.2cm}
\vspace{-0.4cm}

\subsection{CRF system}
Past experiments described in \cite{hahn2011comparing} have shown that the best semantic annotation performance on manual and automatic transcriptions of the MEDIA corpus were obtained with CRF systems. More recently in \cite{vukotic2015time}, this architecture has been compared to popular bi-directionnal RNN (bi-RNN). 
The results was that CRF systems outperform a bi-RNN architecture on the MEDIA corpus, while better results were observed by bi-RNN on the ATIS \cite{hemphill1990atis} corpus. This is probably explained by the fact that MEDIA contains semantic contents whose mentions are more difficult to disambiguate and CRFs make it possible to exploit complex contexts more effectively.

For the sake of comparison with the best SLU system proposed in \cite{hahn2011comparing}, the Wapiti toolkit was used  \cite{lavergne2010practical}. Nevertheless, the set of input features used by the system proposed in this paper  is different from the one used in \cite{hahn2011comparing}.
Among the novelties used in our system, we consider syntactic and ASR confidence features and our configuration template is different. After many experiments performed on DEV, our final feature template includes the previous and following instances for words and POS in a unigram or a bigram to associate a semantic label with the current word. Also associated with the current word are semantic categories of the two previous and two following instances. The other features are only considered at the current position.

Furthermore, the tool discretize4CRF\footnote{https://gforge.inria.fr/projects/discretize4crf/} is used to apply a discretization function to the ASR confidence measures in order to obtain several discrete values that can be accepted as input features by the CRF.

\section{Experimental setup and results}
\label{sec:exp}

Experiments were carried with the MEDIA corpus as in \cite{hahn2011comparing}. For the sake of comparison, the results of their best understanding system is reported in this paper as \textit{baseline}. However, as the WER of the ASR used in this paper is lower (23.5\%) than the one used in the \textit{baseline}, rigorous conclusions can be drawn only on comparisons between the different SLU components introduced in this paper.  

\subsection{The MEDIA corpus} 
\label{sec:media}
%
The MEDIA corpus was collected in the French Media/Evalda project \cite{bonneau2005semantic} and deals with negotiation of tourist services. It contains three sets of telephone human/computer dialogues, namely: a training set (TRAIN) with approximately 17.7k sentences, a development set (DEV) with 1.3k sentences and an evaluation set (TEST) containing 3.5k sentences. The corpus was manually annotated with semantic concepts characterized by a label and its value. Other types of semantic annotations (such as mode or specifiers) are not considered in this paper to be consistent with the experimental results provided in \cite{hahn2011comparing}. Annotations also associate a word sequence to the concepts. These sequences have to be considered as estimations of concept localized mentions. 
Evaluations are performed with the DEV and TEST sets and report concept error rates (CER) for concept labels only and concept-value error rates (CVER)for concept-value pairs.
It is worth mentioning that the number of concepts annotated in a turn has a large variability and may include more than 30 annotated concepts. Among the concepts types there are some, such as three different types of REFERENCE and CONNECTOR of application domain entities. The mentions of these concepts are often short confusable sequences of words.

\subsection{LIUM ASR system dedicated to MEDIA}

For these experiments, a variant of the ASR system developed by LIUM that won the last evaluation campaign on French language has been used~\cite{rousseau2014lium}.
This system is based on the Kaldi speech recognition toolkit~\cite{povey2011kaldi}.
The training set used to estimate the DNN (Deep Neural Networks) acoustic models parameters consists of 145,781 speech segments from several sources: the radiophonic broadcast ESTER~\cite{Galliano06} and ESTER2~\cite{Galliano09} corpora, which accounts for about 100 hours of speech each; the TV broadcast ETAPE corpus~\cite{Gravier12}, accounting for about 30 hours of speech; the TV broadcast REPERE train corpus, accounting for about 35 hours of speech and other LIUM radio and TV broadcast data for about 300 hours of speech. As a total, 565 hours of speech composes the training corpus. These recordings were converted to 8kHz before training the acoustic models in order to be  more appropriate to the MEDIA telephone data.
As inputs, DNN are fed (for training and decoding) with  MFCCs (Mel-Frequency Cepstrum Coefficients) concatenated to $i$-vectors, in order to adapt  acoustic models to speakers.

The vocabulary of the ASR system contains all the words present in the MEDIA training and development corpora, so about 2.5K words.  A first bigram language model (LM) is applied during the decoding process to  generate word-lattices. These lattices are then rescored by applying a 3-gram language model.
In order to get an SLU training corpus close to the test corpus, SLU models are trained with ASR transcriptions. 
To avoid to deal with errors made by an LM over-trained on the MEDIA training corpus, a leave-one-out approach was followed: all the dialogue files in the training  and the development corpora were randomly split into 4 subsets. Each  subset was transcribed by using an LM trained on the manual transcriptions present in the 3 other blocks and linearly interpolated to a 'generic' language model trained on a large set of French newspaper crawled on the web, containing 77 millions of words.
The test data was transcribed with an LM trained on the MEDIA training corpus and the same generic language model.
As shown in table~\ref{tab:wer}, word error rates for the training, development, and test corpora were around 23.5\%.

\vspace{-0.2cm}
\begin{table}[htb]
\centering
\begin{tabular}{|l|c|c|}
  \hline
  train. & dev. & test \\
  \hline
  23.7\% & 23.4\% & 23.6\% \\
  \hline
\end{tabular}
\caption{\label{tab:wer}Word error rates of  transcriptions produced by the ASR systems in order to train, to tune and to test the SLU systems.}
\vspace{-1.0cm}
\end{table} 

\subsection{Results}\label{ssec:results}
Tests were performed for both architectures with the MEDIA DEV set. The best configuration is chosen with respect to the best results observed on the DEV set and applied for obtaining the TEST results. These results in terms of error rate, precision and recall for concepts (C) and concept value (CV) are reported for the best configuration of each architecture in Table \ref{tab:FirstResult}.

\begin{table}[htb]
\vspace{-0.1cm}
\centering

\begin{tabular}{l|c|c|c|c|c|c|}
\cline{2-7}
                               & \multicolumn{3}{c|}{C} & \multicolumn{3}{c|}{CV} \\ \cline{2-7} 
                               & \%Error  & P    & R    & \%Error  & P     & R    \\ \hline
\multicolumn{1}{|l|}{baseline} & 23.8     & -    & -    & 27.3     & -     & -    \\ \hline
\multicolumn{1}{|l|}{NN EDA}   & 22.3     & 0.88 & 0.84 & 28.8     & 0.81  & 0.77 \\ \hline
\multicolumn{1}{|l|}{CRF}      & 19.9     & 0.90 & 0.85 & 25.1     & 0.85  & 0.80 \\ \hline
\end{tabular}

\caption{\label{tab:FirstResult}Comparison on error rate, precision and
 recall for concepts (C) and concept value (CV) pairs
 obtained 
on TEST.}
\vspace{-0.3cm}
\end{table}

It appears that the CRF architecture significantly outperforms NN EDA that shown minor improvements with respect to the baseline.

In order to evaluate the impact of the use of confidence measures among the input features, we made some experiments summarized in Table~\ref{tab:pap-cm}.  As we can see, the confidence measure provided by the MS-MLP  architecture  brings relevant information to reduce the CER and the CVER.

\begin{table}[htb]
\centering
\begin{tabular}{l|c|c|c|c|c|c|}
\cline{2-7}
                             & \multicolumn{2}{l|}{without CM} & \multicolumn{2}{c|}{+pap}& \multicolumn{2}{c|}{+pap +MS-MLP} \\ \cline{2-7} 
                             & C               & CV             & C             & CV    & C             & CV         \\ \hline
\multicolumn{1}{|l|}{CRF}    & 20.9            & 26.0           & 20.5          & 25.7& 19.9          & 25.1          \\ \hline
\end{tabular}
\caption{\label{tab:pap-cm}Impact of the use of confidence measures (probability \textsl{a posteriori} and MS-MLP values) on the performances of CRF on TEST}
\end{table}

\vspace{-0.4cm}
Other versions of the two systems were considered by adding to the usual MEDIA concept labels two more output tags. During training, these tags are replacing the usual one when the hypothesized word is erroneous. If the erroneous hypothesized word is supporting a concept, it is associated to the ERROR-C tag, ERROR-N otherwise. During evaluation, ERROR-C and ERROR-N hypothesized tags are replaced by \textit{null} (tag informing that the word does not convey any MEDIA information) in order to perform the usual MEDIA evaluation protocol. Results on TEST, obtained with the best configuration observed on DEV, are reported in Table \ref{tab:SecondResult}.

\begin{table}[htb]
\centering

\begin{tabular}{l|c|c|c|c|c|c|}
\cline{2-7}
                             & \multicolumn{3}{c|}{C}         & \multicolumn{3}{c|}{CV}        \\ \cline{2-7} 
                             & \%Error & P             & R    & \%Error & P             & R    \\ \hline
\multicolumn{1}{|l|}{NN EDA} & 22.1    & 0.90          & 0.82 & 27.8    & 0.84          & 0.77 \\ \hline
\multicolumn{1}{|l|}{CRF}    & 20.6    & \textbf{0.91} & 0.84 & 25.4    & \textbf{0.86} & 0.79 \\ \hline
\end{tabular}

\caption{\label{tab:SecondResult}Evaluation results on TEST obtained by adding to the usual MEDIA concept label the ASR error detection tags ERROR-C and ERROR-N.}
\vspace{-0.4cm}
\end{table}

Results in Table \ref{tab:SecondResult} are similar to those in Table \ref{tab:FirstResult}, but we can notice some small differences. For instance,  precision is now better, even if the CER is not reduced for CRF while it is for NN-EDA.  
Using these four SLU systems that can be executed in parallel, it is worth trying to see if improvements can be obtained by their combination with weight estimated by optimal performance on the DEV set. The results are reported in Table \ref{tab:ConsensusResult} and compared with the ROVER \cite{fiscus1997post} combination applied to the six SLU systems described in \cite{hahn2011comparing}.


\begin{table}[htb]
\centering

\begin{tabular}{l|c|c|c|c|c|c|}
\cline{2-7}
                                           & \multicolumn{3}{c|}{C}               & \multicolumn{3}{c|}{CV}              \\ \cline{2-7} 
                                           & \%Err.       & P             & R    & \%Err.       & P             & R    \\ \hline
\multicolumn{1}{|l|}{baseline comb.} & 23.1          & -             & -    & 27.0          & -             & -    \\ \hline
\multicolumn{1}{|l|}{CRF+NN comb.}   & \textbf{19.3} & 0.91          & 0.85 & \textbf{24.5} & 0.86          & 0.80 \\ \hline
\multicolumn{1}{|l|}{CRF+NN cons.}     & -             & \textbf{0.96} & 0.72 &               & \textbf{0.89} & 0.68 \\ \hline
\end{tabular}

\caption{\label{tab:ConsensusResult}Performance of the weighted combination or the consensus of outputs from the four systems on TEST.}
\vspace{-1.0cm}
\end{table}

The results show 0.6\% and 0.6\% absolute reductions for CER and CVER with respect to the best CRF architecture and 4.5\% and 2.8\% with respect to the baseline. Considering that the best results on manual transcriptions are above 10\% on the TEST set, one may conclude that, with the solutions presented in this paper, the contribution of ASR errors to the overall SLU errors is inferior to errors observed for manual transcriptions.
A detailed analysis of the errors observed in the automatic and manual transcriptions show a common large error contributions for concepts such as three different types of reference, connectors between domain relevant entities, and proper names that can be values of different attributes. These concepts are expressed by confusable words whose disambiguation requires 
complex context relations that cannot be automatically characterized (at least with the available amount of train data) by CRFs nor by the type of attention mechanisms used in NN EDA.

Considering the case in which all the four systems provided the same output (consensus) for each word, a 0.96 precision with 0.72 recall were observed on the TEST set. Lack of consensus in the DEV and the TEST sets appears to correspond in most cases to mentions of only few types of concepts. This is a very interesting result since it suggests that further investigation on these particular cases is an important challenge for future work.





  
  






\section{Conclusions}

Two variations of two SLU architectures respectively based on CRFs and NN-EDA have been considered.  Using the MEDIA corpus, they were compared with the CRF SLU, considered as baseline that provided the best results among seven different approaches as reported in \cite{hahn2011comparing}. The main novelties of the proposed SLU architectures are the use, among others, of semantically relevant confidence and input features. The CRF architectures outperformed the NN-EDA architectures with significant improvement over the baseline. Nevertheless, NN-EDA architectures appeared to be useful when combined with the CRF ones. The results show that the interaction between the ASR and SLU components is beneficial. Furthermore, all the architectures show that most of the errors are for concepts whose mentions are made of short confusable sequences of words that remain ambiguous even if they can be localized. These concept types are difficult to detect, even on manual transcription, indicating that the interpretation of the MEDIA corpus is particularly difficult. Thus, suggested directions for future work should consider new structured mechanisms of attention capable of selecting features of distant contexts in a conversation history. The objective is to identify a sufficient set of context features for disambiguating local concept mentions.



\bibliographystyle{IEEEtran}

\bibliography{mybib}

\end{document}